\documentclass{aamas2010_cameraReady}

\DeclareMathSizes{9}{8.5}{6}{6}   

\begin{document}



\AuthorsForCitationInfo{Mark P. Woodward and Robert J. Wood}

\TitleForCitationInfo{Learning from Humans as an I-POMDP}

\title{Learning from Humans as an I-POMDP}



%
%
%
%

%

\numberofauthors{2}

\author{
%
\alignauthor
Mark P. Woodward\\
       \affaddr{Harvard Micro-robotics Laboratory}\\
       \affaddr{60 Oxford Street}\\
       \affaddr{Cambridge, Massachusetts, 02138}\\
       \email{mwoodward@eecs.harvard.edu}
\alignauthor
Robert J. Wood\\
       \affaddr{Harvard Micro-robotics Laboratory}\\
       \affaddr{60 Oxford Street}\\
       \affaddr{Cambridge, Massachusetts, 02138}\\
       \email{rjwood@eecs.harvard.edu}
}

\maketitle

\begin{abstract}
The interactive partially observable Markov decision process
(I-POMDP) is a recently developed framework which extends the POMDP
to the multi-agent setting by including agent models in the state
space. This paper argues for formulating the  problem of an agent
learning interactively from a human teacher as an I-POMDP,
where the agent \emph{programming} to be learned is captured by random
variables in the agent's state space, all \emph{signals} from the
human teacher are treated as observed random variables, and the human
teacher, modeled as a distinct agent, is explicitly represented in the
agent's state space.
The main benefits of this approach are: 
i. a principled action selection mechanism, 
ii. a principled belief update mechanism, 
iii. support for the most common teacher \emph{signals}, and 
iv. the anticipated production of complex beneficial interactions.
The proposed formulation, its benefits, and several open questions are
presented.

\end{abstract}

\section{Introduction}

We propose formulating the problem of learning interactively from a
human teacher as an interactive partially observable Markov decision
process (I-POMDP). The I-POMDP is a formulation of the decision making
problem faced by an agent in a stochastic, partially observable,
multi-agent environment \cite{Gmytrasiewicz:2005}. In our formulation
the human teacher is modeled as a distinct agent in the agent's state 
space. The parameters of the agent programming to be learned are 
expressed as random variables also in the agent's state space.
Finally, all signals from the human teacher are represented as
observed random variables.

Some of the benefits of this approach are a principled action
selection mechanism, a principled belief update mechanism, the
foreseen production of complex interactions with the human teacher,
a single mechanism for interpreting all teacher signals including
gestures, body posture, natural language, and direct modeling, and a
single mechanism for selecting all agent actions including emotive
displays, world manipulation, and spoken language.

In the following sections we present the proposed formulation, discuss
its benefits, and finish with several open questions. But first we
begin with a brief description of the envisioned domain for the
our formulation.

\section{Domain}

The domain envisioned for this paper is an agent learning
interactively from a non-technical human teacher. The interaction 
consists of signals generated by the teacher and the
agent. Some examples of these signals are words, gestures, facial
expressions, body posture, eye gaze, and rewards. We intend the proposed
formulation to cover most forms of teaching including learning from
demonstration and learning from reinforcements.

An important restriction of our proposal is that a teacher must be
present for learning to take place. The teacher is explicitly modeled
by the agent and learning is accomplished by interpreting signals
resulting from the interaction with the teacher.\footnote{This does
not preclude sub-systems from performing \emph{independent} learning;
for example, in a robotics domain, the agent might independently learn
through experience that bumping into walls is disadvantageous.} The
main reason for this restriction is that the agent is seen as
attempting to learn what the teacher wants it to learn, and, in general,
the teacher needs to be present to convey what they want learned.

\section{Proposed Framework}

We briefly summarize the partially observable Markov decision process
(POMDP) and its extension, the interactive partially observable
Markov decision process (I-POMDP), before describing our formulation of
learning from a human teacher as an I-POMDP.

\subsection{POMDP}
A partially observable Markov decision process is a formulation of an
agent's decision process when operating in a sequential, stochastic,
partially observable domain. Essentially, the agent knows only
probabilistically how the world changes around it and only
probabilistically how its sensor readings reflect the state of
the world. Importantly, the optimal decision (a.k.a action) is the one that
maximizes the agent's expected utility. It is an
\emph{expected} utility because the agent knows only probabilistically
the current state of the world, how it's actions will affect the
world, and what measurements it will receive assuming the world gets
to a certain state. A POMDP is captured by the tuple $\langle S,A,T,
\Omega,O,R \rangle$. $S$ is the set of world states, $A$ is the set of
actions the agent can perform, $T$ is the motion model
defining the probability of reaching any state $s' \in S$ given an action
$a \in A$ executed from a state $s \in S$, $\Omega$ is the set of measurements
(a.k.a observations) the agent might receive, $O$ is the measurement
model (a.k.a observation model) defining the probability of measuring
$o \in \Omega$ given the world is in state $s \in S$, and finally $R$ is the
utility function (a.k.a reward function) mapping states of the world
or sometimes belief states to real numbers. It is some function of
$R$ that the agent is trying to maximize when selecting the next
action, for example the discounted sum of $R_t$ from $t=0$ to some
horizon $t=T$. 

In the case of discrete states, the recursive Bayesian belief update for time step $t$
having taken action $a_{t-1}$ and received measurements $o_t$ is:\footnote{The continuous case is similar
with integrals replacing summations and pdfs in place of pmfs for the
beliefs $b_t$.}
\begin{equation}
  b_t(s_t) = \beta O(o_t|s_t)\sum_{s_{t-1}}T(s_t|s_{t-1},a_{t-1})b_{t-1}(s_{t-1}),
\end{equation}
where $\beta$ is the normalizing constant, 

Action selection based on maximum expected utility is defined as
\begin{equation}
\arg\max_{a_t}EU_t(b_t)
\end{equation}
where 

\begin{equation}
\begin{split}
EU&_t(b_t) = \\
&R(b_t) + \gamma\max_{a_t}\sum_{o_{t+1}}P(o_{t+1}|b_t,a_t)EU_{t+1}(b_{t+1}|a_t,o_{t+1}),
\end{split}
\end{equation}
$\gamma$ is the discount factor, and
\begin{equation}
P(o_{t+1}|b_t,a_t) = \sum_{s_{t+1}}O(o_{t+1}|s_{t+1})\sum_{s_t}T(s_{t+1}|s_t,a_t)b_t(s_t).
\end{equation}
For finite horizon solutions with horizon {T}, this recursion
terminates with 
\begin{equation}
EU_T(b_T)=R(b_T), 
\end{equation}
where $R$ is the specified utility function. As an example of $R$,
in our initial robot learning experiments, we have used $R(b_T) =
-Entropy(b_T)$, Since entropy is a measure of uncertainty in the
distribution, using negative entropy as the utility function means
that the robot chooses actions that maximize certainty over the
parameters being learned. 

The above action selection equations form a decision tree into the
future, where the tree nodes are belief states, and the branches alternate
between measurements and actions.

For further details on POMDPs see \cite{Kaelbling:1998} and \cite{Thrun:2005}.

\subsection{I-POMDP}
An I-POMDP extends the POMDP to the multi-agent setting, for our case
there will only be a single other agent which is the human teacher.
An I-POMDP of an agent $i$ is captured by the tuple $\langle IS_i,A,T_i,
\Omega_i,O_i,R_i \rangle$, which is the usual POMDP tuple, except
$IS_i$ includes a model of the other agents, which could themselves be
I-POMDPs, and $A$ is extended to include all actions that all agents 
could take. If the teacher is modeled with an I-POMDP, we might also
model the teachers model of the agent as another I-POMDP. In order to
reach a solution, this recursive agent model nesting must eventually
ground out with a non interactive model, such as a POMDP. 

Belief updates and action selections are similar to the POMDP case,
except that measurements, actions, and beliefs of the teacher, and of
the teacher about the agent, etc. need to be incorporated. We leave
these details to the referenced readings. The above POMDP equations
give a good intuition for the types of computations required in an
I-POMDP.

For further details on I-POMDPs see \cite{Gmytrasiewicz:2005}, it may
also be helpful to read their earlier work on nested models as a
primer \cite{Gmytrasiewicz:2000}.

\subsection{Learning from Humans as an I-POMDP}
In our proposed formulation, the agent programming is defined by
random variables stored in the agent's state space $S$ within
$IS$. For example in a work-flow, these parameters could be the trigger
conditions, the number of nodes, and the value of transitions.
Another example could be the parameters defining the angle of a
driver when engaging with a screw. 
Also in our proposal, all signals in the human robot interaction are
expressed as measurements $o \in O$. Examples of signals were given in
the Domain section above.
Lastly, the human teacher is explicitly represented as an interactive
agent in the agent's state space. This means that the
teacher's belief about the state of the world, about the parameters of
the programming, and about the belief of the agent are
maintained and updated by the agent. Additionally, the 
agent, when selecting actions to maximize it's expected utility, can
take into account likely future actions of the teacher, and its own
responses to those future actions, resulting in a potentially high
utility state. Most of the interactions listed below in the Benefits
section are a result of this recursively nested modeling.

On each
time-step, the agent first updates its beliefs and then selects an
action to perform. The beliefs being updated include the agent's
belief about the parameter's of the programming and the physical state
of the world, the agent's belief about the action performed by the
teacher on the previous time-step, the agent's belief about the
teacher's belief about the parameters of the task and the physical
state of the world, the agent's belief about the teacher's belief
about the agent's previous action, etc. 

With the belief state updated, the agent then selects an action. One
method to select this action would be, for each nested belief level,
to search out to a time horizon $T$ over all possible actions and
measurements which could be performed or received by both agents,
evaluate the utility of the resulting belief state at the horizon $T$
and then work this utility back to the current actions by taking
expectation at measurement branches and maximization at action
branches. Optimal actions in lower nesting levels appear to the higher
level agent as distributions over actions by the lower level agent.

The selected action would then be performed, and along with the actions
of the teacher, would affect the state of the world, which would, on
the next time-step, be perceived through measurements by the teacher
and the agent. And the process would repeat.

\subsection{A Note on Complexity}

The computation required for the belief update and action selection
mentioned in the previous time grows exponentially with depth of
agent model nesting, the time-steps to the horizon, and the complexity of the
state space. We believe that sampling techniques will likely be the
the best solution to this exponential growth. Particle filters
have been applied to exponential state growth \cite{Fox:1999}
and to exponential nested agent models \cite{Doshi:2005} to
good result, but we are unaware of effective techniques to
handle exponential growth due to action and measurement branching in
planning over future time-steps. Hopefully the exponential growth due
to action branches can be moderated by sampling over actions based on
heuristics learned over time. Similarly the exponential growth due to
measurement branches might be moderated by sampling according to the
likelihood of measurements. 

Even with the proposed sampling techniques, computation will remain a
major problem with the proposed formulation. We believe this is a
necessary side effect of the complexity of the beliefs and plans being
represented and that the benefits of this complexity, described in the
next section, warrant wrestling with the exponential growth of
computation. 

\subsection{A Note on Modeling}

The presented formulation is model based, in that it requires a
probabilistic motion model for the state space and probabilistic measurement
models for the measurements observed by the agent. Additionally, the
recursive agent models must ground out with a stochastic model of
actions for either the teacher or agent. All of these models
have parameters that need to be set, for example the attention
span of a typical student, the rate of misspoken words by a human
teacher, or the variance of a laser range scan. As the proposed
technique matures, future agents will likely maintain these parameters
by long term monitoring of their physical sensors and repeated
interactions with humans. In the near term, these parameters can be set by
controlled sensor calibration and data and from experiments with a
human teaching another human.

\section{Benefits}

In this section we describe many of the foreseen benefits of the
proposed formulation. We begin with benefits to the learning from
human research community due to the formulation being based on
recognized principles. We then describe some complex interactions that
should result from the formulation. Followed by a discussion of the
extendability of the formulation to other interactive agent
domains. We then finish with some miscellaneous benefits of the
proposed formulation.

\subsection{Principled Formulation}

One of the main benefits of the proposed formulation is that it is
based on recognized principles. Firstly, the belief over the state of the
world, the parameters of the programming to be learned, and models of
the teacher are all updated using Bayesian Inference, with the
resulting implication that \emph{all} signals by either agent be expressed as
observed random variables. Secondly, the agent's actions are selected
in accordance with the principle of maximum expected utility. By
formulating the problem of learning from humans based on the two
principles of Bayesian Inference and maximum expected utility,
techniques we develop can be useful to, and we can make use of
techniques from, many fields which commonly base techniques on these
principles, such as the broader reinforcement learning community,
signal processing, and operations research.

\subsection{Complex Interactions}

As a product of the nesting of agent beliefs (beliefs about the teacher's
beliefs about the agent's beliefs etc.), we expect to see a number of
interesting exchanges. One type of exchange is where the agent
has an inconsistency in its nested agent models and determines that
acting to reduce this inconsistency will result in the highest expected
utility.\footnote{The examples assume a utility function that favors
faster learning, for example maximizing the discounted negative
expected utility of the random variables defining the programming.}
The following are three examples of the agent taking this type of
action to correct an inconsistency.

\begin{itemize}

\item
\textbf{Interruption:} The teacher is teaching about $x$, and the
agent interrupts to inform the teacher that they are clear on $x$, but
more uncertain on $y$.

\item
\textbf{Clarification:} The agent interrupts the teacher to clarify
that a previous action came through as $x$, and questions did they
actually intend $y$.

\item
\textbf{Correction:} This is the inverse of clarification, the agent
interrupts the teacher to communicate that the agent believes that the
teacher believes that the agent was asking about $x$, when it was actually
asking about $y$.

\end{itemize}

I-POMDPs have been demonstrated to generate similar exchanges in
simple cooperative multi-agent settings \cite{Gmytrasiewicz:2001}.

\subsection{Extendability}

Another benefit of the proposed formulation is that it should easily
extend to other useful agent settings. For example, in the proposed
approach, as a sub-task of action selection, the agent is already
determining optimal actions for the teacher, 
just one nesting level down, thus it should be straightforward to
extend the formulation to the agent teaching a human or another
agent. Similarly, the agent should be able to interactively learn from
a non-human teaching agent. Additionally, since I-POMDPs were developed for the
multi-agent setting, our formulation should extend to the common
multi-agent settings of cooperation and competition, be it with human
or non-human agents.

\subsection{Miscellaneous Benefits}

This section describes benefits that do not fall under the previous
categories but are worth mentioning. 

The first is that all actions fall under one umbrella. This means that
we do not need separate action selection systems for physical
movement, words, text, beeps, facial expressions, etc. As mentioned
above, the agent will likely have built-in heuristics for determining
the utility of specific actions to speed up action
selection.\footnote{The proposed approach of a unified action
selection mechanism is in stark contrast to the distributed
behavior based robotics approach of \cite{Brooks:1991b} which is
likely a more accurate model of human action selection. We feel that
having an identifiable, and thus easily adjustable, object function
and action selection mechanism justify deviation from a more accurate
distributed human action selection model which may be harder to
develop.}

Secondly, a number of action selections which may have needed explicit
coding under other formulations happen without coding under the
proposed formulation, because they fall out as the rational
action selection. The following are several examples of behavior that
should be exhibited without coding:

\begin{itemize}

\item Facial expressions, such as raising eyebrows for a social robot,
  become the rational action, perhaps because conveying uncertainty
  will modify the teachers belief about the robots belief about the
  task in such a way that the teacher will provide further instruction
  resulting in a higher expected utility.

\item Similarly, periodically looking at the teacher might be rational
  behavior since it could maintain the teacher's belief that the agent
  is still focused on learning, avoiding the potentially time
  consuming and consequently low utility producing actions by the
  teacher of checking if the agent is still paying attention.

\item Actions will be intuitively ordered, for example if a robot
  agent is going to ask the teacher about a series of objects and
  moving to each object takes time, the rational behavior would be
  to ask about objects in an order that minimizes travel time;
  The faster questions are asked and answered the higher the expected
  utility should be.

\item Often not acting will be the optimal action, since
  actions may interrupt the teacher, who is providing useful
  information, resulting in a lower expected utility than performing
  no action at all.

\item In a robot agent, the agent may perform pointing actions to
  direct the teacher's attention. Both the decision to point and the
  duration of the point would be handled automatically by the
  system. The agent would compute the expected utility of various
  pointing durations and choose the duration with the maximum
  utility. A short point would allow the agent to move on but would
  leave the agent uncertain about the teacher's attention, with a long
  point the agent would be certain about the teacher's attention but
  will have pushed higher utility states further into the
  future. The principle of maximum expected utility would give the
  optimal duration to use. 

\end{itemize}

\section{Questions}

In this section we first answer a couple questions posed by the
organizers, before posing several open ended questions of our own.

\subsection{Organizer Questions}

\begin{description}

\item[Question:]
  How explicitly pedagogical is the human teacher?  At one extreme,
  the human is a role-model not considering the agent at all; at the
  other, the human is carefully formulating a curriculum.

\item[Answer:]
  Our approach is towards the pedagogical end of the spectrum. An
  intelligent agent, explicitly instructing the robot is
  required. Though the signals of instruction can be ambiguous and
  varied (demonstration, modeling, reinforcement, or natural
  language).

\item[Question:]
  How dependent is learning on communication between human teacher and
  agent?  At one extreme, the human and agent merely observe one another
  interacting with the environment; at the other there is a complex
  dialog between teacher and student.

\item[Answer:]
  The proposed formulation assumes a dialog of signals between the
  teacher and the agent. That said, the formulation should extend to
  the learning by observing peer agents. Since other agents in the
  environment are assumed to be rational, observing their actions
  updates the belief about the random variables affecting their
  actions. Through observation, the belief about their objective
  function is also updated. If this objective function is deemed
  ``similar" to the agent's, then imitation is reasonably the rational
  behavior that would result.

\item[Question:]
  How interdependent is learned information?  At one extreme, learning
  can potentially happen in any order (e.g., mapping a state space); at
  the other, each new piece of knowledge must be formulated in terms of
  the previous one (e.g., Kirchhoff's laws depend on current and voltage).

\item[Answer:]
  Learned information is not inherently hierarchical in our
  approach. But, for example, it is likely rational for steps of a
  task to be taught in order, or reverse order, and since the robot
  assumes a rational teacher, the learning task would be much
  harder if the teacher choose to teach step one, then five, then two,
  as apposed to one, then two, then three, or three, then two, then
  one. What is rational depends on the zeroth level models of the
  teacher and robot. If these zeroth models have a certain flow to
  them, then the agent will learn best under that flow.

  Also, making use of prior learned programming as heuristics for
  future learning is likely a good idea. For example, learning the
  task of clearing a table should be easier having learned to set the
  table; forks and spoons moved together before, so they will likely
  move together again.

\item[Question:]
  What is the relative importance of how learning occurs vs. the end
  result of learning in the research?  At one extreme, learning from
  humans is just a pragmatic way of configuring a real-world system; at
  the other, the system is only valued for the insight it provides on
  human learning.

\item[Answer:]
  We view learning from humans as a pragmatic way of configuring a
  real-world system. 

  Notably, we view our approach as a poor approximation to human
  learning. Through introspection, we may think that our actions are
  chosen by maximizing our expected utility, but psychology studies
  have shown that that humans often do not select actions
  to maximize their expected utility\cite{Kahneman:1971}.
  And (Brooks 1991) makes a compelling argument that model based
  planning is not how biological agents select
  actions\cite{Brooks:1991b}.

  As noted above, our approach is attractive, not because it resembles
  human learning, but because it is a principled approach to robot
  behavior during learning, in addition to the other benefits
  mentioned.

\end{description}

\addtolength{\textheight}{-4.5in} 
\subsection{Open Questions}

\begin{itemize}
\item
   What is the \emph{right} objective function? If pleasing the human, how do we
     measure this?
\item
   What are reasonable metrics for evaluating this formulation?
   e.g. time to teach, certainty/accuracy vs. time curves, ease of
   use.
\item
   At what granularity if any, does this become less ``effective"
     than model free techniques such as Q-Learning?
     than model based techniques such as Value Iteration?
\item
   Are there any situations in which a signal from the human teacher
   should be considered part of the utility function? 
\item
   How deep should the agent belief nesting be?
\item
   What should the ground agent model be?
\item
   What is the simplest student/teacher game to explore this formulation?
\item
   Since computation is an issue, how should we trade off accuracy of
   belief representation, depth of agent model nesting, and depth of
   search?
\item
   How can we make use of heuristics?
\item
   How can these heuristics be learned?
\end{itemize}

\section{Conclusions}

This paper proposed formulating the problem of learning from humans as
an interactive partially observable Markov decision process (I-POMDP),
where parameters defining the agent programming to be learned are
captured as random variables in the agent's state space and the
teacher is explicitly modeled as a distinct agent also in the agent's
state space.
POMDPs and I-POMDPs were briefly described before outlining the 
proposed formulation. 
We described several benefits of the
formulation, namely principled action selection, principled belief
updates, and the anticipated production complex interactions. 
We finished by posing several open questions regarding the formulation.

We believe that the benefits of the new formulation justify its
computational complexity, and will help to advance research on
agents learning from human teachers.

%
\bibliographystyle{abbrv}

\bibliography{woodward}  

\begin{thebibliography}{1}

\bibitem{Brooks:1991b}
R.~A. Brooks.
\newblock Intelligence without reason.
\newblock In {\em Proc.~International Joint Conference on Artificial
  Intelligence (IJCAI)}, 1991.

\bibitem{Doshi:2005}
P.~Doshi and P.~J. Gmytrasiewicz.
\newblock Approximating state estimation in multiagent settings using particle
  filters.
\newblock In {\em Proc.~Internation Conference on Autonomous Agents and
  Multiagent Systems (AAMAS)}, 2005.

\bibitem{Fox:1999}
D.~Fox, W.~Burgard, F.~Dellaert, and S.~Thrun.
\newblock Monte carlo localization: Efficient position estimation for mobile
  robots.
\newblock In {\em Proc.~of the National Conference on Artificial Intelligence
  (AAAI)}, 1999.

\bibitem{Gmytrasiewicz:2005}
P.~J. Gmytrasiewicz and P.~Doshi.
\newblock A framework for sequential planning in multi-agent settings.
\newblock {\em Journal of Artificial Intelligence Research}, 24:49--79, 2005.

\bibitem{Gmytrasiewicz:2000}
P.~J. Gmytrasiewicz and E.~H. Durfee.
\newblock Rational coordination in multi-agent environments.
\newblock {\em Autonomous Agents and Multi-Agent Systems Journal}, 3:319--350,
  2000.

\bibitem{Gmytrasiewicz:2001}
P.~J. Gmytrasiewicz and E.~H. Durfee.
\newblock Rational communication in multi-agent environments.
\newblock {\em Autonomous Agents and Multi-Agent Systems Journal}, 4:233--272,
  2001.

\bibitem{Kaelbling:1998}
L.~P. Kaelbling, M.~L. Littman, and A.~R. Cassandra.
\newblock Planning and acting in partially observable stochastic domains.
\newblock {\em Artificial Intelligence}, 101:99--134, 1998.

\bibitem{Kahneman:1971}
D.~Kahneman and A.~Tversky.
\newblock Subjective probability: A judgment of representativeness.
\newblock {\em Cognitive Psychology}, 3:430--454, 1971.

\bibitem{Thrun:2005}
S.~Thrun, W.~Burgard, and D.~Fox.
\newblock {\em Probabilistic Robotics}.
\newblock MIT Press, Cambridge, MA, 2005.

\end{thebibliography}
%
%
\end{document}